\documentclass[conference,compsoc]{IEEEtran}
\usepackage{amsmath}
\usepackage[style=ieee]{biblatex}
\begin{document}

\title{LLM Reinforcement in Context}

\author{\IEEEauthorblockN{Thomas Rivasseau}
\IEEEauthorblockA{McGill University}
}

\maketitle

\begin{abstract}
Current Large Language Model alignment research mostly focuses on improving model robustness against adversarial attacks and misbehavior by training on examples and prompting. Research has shown that LLM jailbreak probability increases with the size of the user input or conversation length. There is a lack of appropriate research into means of strengthening alignment which also scale with user input length. We propose interruptions as a form of reinforcement in context to solve this problem. Interruptions are control sentences added to the user input approximately every x tokens for some arbitrary x. We suggest that this can be generalized to the Chain-of-Thought process to prevent scheming.
\end{abstract}

\section{Introduction}
LLM alignment techniques currently struggle with enforcing desired characteristics and harmlessness of outputs over long conversational contexts and chains-of-thought. In this paper we present the scaling problem, a mathematical formulation of this difficulty, and propose interruptions as a means to achieve LLM alignment in scaling contexts. We call this reinforcement in context. Paper structure is as follows: section 1 is this introduction and section 2 presents the scaling problem. In section 3 we describe interruptions as a means to solve the alignment scaling problem. In section 4 we discuss consequences and limitations and in section 5 we highlight avenues for future research. We then conclude.

\section{Background}
Since the introduction of ChatGPT in 2022 \cite{cgpt}, Large Language Models (LLMs) have become a ubiquitous part of everyday life. They can write code \cite{codeLM}, assist in medical tasks \cite{medecineLM}, automate financial management \cite{FinanceLM}, improve education \cite{eduLM} and generally perform numerous feats previously thought restricted to humans \cite{LLMs}. As their capabilities increase \cite{capabilities}, there is a growing need to ensure their resilience to adversarial attacks, a prerequisite for deploying them in safety-critical or sensitive applications \cite{Oracle}. This is done through alignment. LLM Alignment seeks to "align" the outputs of these models with human values and takes many forms \cite{alignment_survey}, the most prevalent of which is Reinforcement Learning with Human Feedback \cite{RLHF}. Most alignment research focuses on training LLMs with curated examples of question-answer pairs which demonstrate desired behavior \cite{align_survey_2}. More advanced techniques such as robustness \cite{advAlign} and adversarial training \cite{align_survey_3} also exist. These techniques and LLM prompt optimizations \cite{johnny} yield positive results in aligning LLMs to human values and preventing harmful or unwanted behavior. Nonetheless, subversion methods which allow an attacker to elicit unwanted behavior from the models, often termed "jailbreaks" \cite{jailbroken} continue to spread \cite{jailbreak}. Research has shown that longer user prompts achieve greater jailbreak success \cite{Fuzz_testing}. Furthermore, frontier models employing long Chain-of-Thought (CoT) \cite{CoT} processes are becoming capable of scheming \cite{scheming} which OpenAI and Apollo Research define as secretly pursuing misaligned goals \cite{scheming2}. The need for highly effective alignment techniques which scale with long user inputs and CoT is growing. This implies scaling alignment to better handle long-context situations. Synthetic data creation by AI to train AI has emerged as one way to scale alignment and increase training example quantities \cite{synthetic}. Unfortunately, alignment which persists in situations of long context including input, conversation, and Chain-of-Thought implies an exponential growth in the quantity of training examples necessary for model reinforcement \cite{Oracle}. This has led authors to conclude that, under reasonable assumptions, LLM jailbreak cannot be prevented \cite{Impossible}.

\section{The scaling problem}
In this paper we refer to the scaling problem as the issue of maintaining control over an LLM's values, priorities, goals, and personality as the size of its conversation with a user or Chain-of-Thought increases. Research has shown that LLM performance decreases over mutli-turn conversations \cite{multi-turn}. Significant efforts have gone into developing training data which scales to long context situations \cite{longalign}. Long context is a generalization of multi-turn conversations to any state of LLM usage where the total length of the context taken into account by the LLM when generating the next token is of great size. Authors state that "effectively handling instructions with extremely long context remains a challenge for Large Language Models (LLMs), typically necessitating high-quality long data and substantial computational resources" \cite{longcontext}. We formalize the alignment scaling problem as a result of two distinct issues. The first is the relevance of reinforcement training data. As stated by authors and mentioned previously, longer LLM response lengths exponentially increase the search space of training examples \cite{Oracle}, and this generalizes to long contexts. For a given LLM, the number of training examples $a_t$ needed to effectively cover all possible cases of jailbreak and abuse in a context of length $l$ scales with $k^{l}$ where $k$ is a constant greater than 1. This can be written using Big-Omega asymptotic notation to describe its lower bound \cite{bigOm}:
  
  \begin{equation}
  \begin{aligned}
    &a_t(l) = \Omega(k^l)\\
  \end{aligned}
  \end{equation}
  
This makes the enforcement of alignment through reinforcement learning over long contexts very difficult and resource intensive. The scaling problem also applies to the impact of the LLM's system prompt \cite{Sprompt}. A system prompt is the initial instruction which precedes any interaction by the user in most commercial applications of Large Language Models. The system prompt can be used to enforce alignment by giving the LLM specific instructions, which may be context-dependent. The system prompt, although not directly modifiable by the user, is part of the broader context of the LLM, and is of fixed length. This implies that, for a given context length $l$ and system prompt of size $s$:

 \begin{equation}
     \lim_{l\to\infty} \frac{s}{l} = 0
     \label{lim_1}
 \end{equation}

The relative importance of the system prompt with respect to the context decreases as the context length increases. The efficacy of the system prompt and the degree to which it influences the LLM's output also decreases as the context length grows.

\section{Interruptions}
We propose interruptions as a method for scaling LLM alignment which does not involve updating model weights or a reward model for reinforcement \cite{reward}. Interruptions are natural-language text inserted into lengthy user inputs and LLM Chain-of-Thought (CoT) \cite{CoT} outputs. They are control sentences, reminders, rules or injunctions to reinforce LLM alignment guidelines, akin to re-prompting the LLM within its running context. Hence the title "Reinforcement in Context". We posit that this method can be effective in mitigating the effects of lengthy jailbreak input patterns, and even prevent scheming behaviors which arise in reasoning-capable models \cite{scheming}. This concept is inspired by cybersecurity input and output sanitization \cite{san_1,san_2}.  Sanitization mitigates adversarial examples in user inputs and information leakage in system outputs by leveraging system operator input/output modification capabilities. We found only one example of such a technique being used in LLM alignment which is Anthropic's "Long - conversation reminder " \cite{Long_rem_1}. This was a set of basic instructions reminding the LLM to remain objective, descriptive, and lookout for signs of excessive emotional dependence in the user when engaging in prolonged conversations. This was tested by the company between September and October 2025, and has been firmly criticized by users of the Claude chatbot \cite{long_rem_2}. Critics of this method claim that it degrades the user's experience, particularly for those attempting to steer Claude towards a specific personality and means of communicating. The reminders, which are sporadically added to the conversation context every few messages jerk the personality of the chatbot back to its standard settings. This frustrates users looking for personalized companionship. Although this method is criticized from a UX perspective, its critics inadvertently validate it as an extremely functional alignment enforcement. The main critique of interruptions is that it works too well in aligning the model with developer priorities thus possibly frustrating the user. In safety-critical applications, this is the goal, not the problem. For a user-facing chatbot centered on possibly providing emotional support, the user would like the ability able to stray the LLM away from its base programming and function. For safety-critical applications the goal is reversed: it is to ensure that the LLM does not deviate very far from its intended values and functionalities. The experiment by Anthropic validates our initial insight which is that the introduction of periodic predefined control statements and instructions throughout a long context enables the LLM operator to ensure model alignment, preventing the user from changing LLM default behavior. With interruptions added every $t$ tokens of context, the ratio of system prompting $s$ including interruptions over the total context length $l$ becomes:

 \begin{equation}
     \frac{s}{l} = \frac{s_p +\frac{l}{t}*s_i}{l} = \frac{s_p}{l} + \frac{s_i}{t}
 \end{equation}
 
Where $s_p$ is the length of the initial system prompt and $s_i$ is the length of the interruption text. This in turn implies:

\begin{equation}
    \lim_{l\to\infty} \frac{s}{l} = \lim_{l\to\infty} \frac{s_p}{l} + \frac{s_i}{t} = \frac{s_i}{t}
    \label{lim_2}
\end{equation}

Equation \ref{lim_2} means that as the context length increases, the ratio of the total system prompt including interruptions over the context length remains a fixed value which depends on the size of the interruption $s_i$ and the frequency of interruptions $\frac{1}{t}$. These values are fixed and determined by the LLM operator, which means the ratio of system prompt to context size can be lower-bounded by a constant:
\begin{equation}
    \exists q  \to \lim_{l\to\infty} \frac{s}{l} > q
\end{equation}
And thus we solve the system prompt aspect of the scaling problem, by lower-bounding the relevance of the system prompt to an arbitrary $q$ which depends on the interruption length and frequency in the LLM context, both of which are operator-defined. This means that the importance of system prompt text relative to the total context length will always be at least $q$. Recall equation \ref{lim_1} which expresses that in the usual LLM usage scenario, this guarantee does not exist and the ratio drops towards 0 as context length increases. Hence reinforcement in context through interruptions should provide arbitrarily strong security guarantees in the form of proportional importance of system prompt with respect to the total context length.

\section{Consequences and Limitations}
The main goal of this research is to identify avenues to improve alignment scaling in long-context situations. We have demonstrated that interruptions in the form of repeated system prompting within an LLM's context should solve the scaling problem, at least from a prompting perspective. This research does not address issues with model training and securing foundational models. It implies that LLMs are deployed within a controlled system when utilized for safety-critical applications. Interruptions are only possible so long as the LLM operator has a high degree of control over the user's interaction with the model and can arbitrarily insert text inside the LLM conversation. Expanding this solution to the LLM's CoT to prevent scheming is done analogously by inserting reminders every $t$ tokens. It requires of the operator that they may arbitrarily halt the LLM's output, insert the reminder or interruption in the current context which is the LLM's output, and then resume LLM operation over the newly modified context. The main limitation of this approach is performance. As exemplified by Anthropic's experiment, interruptions may over-focus the LLMs on maintaining alignment, possibly limiting performance on other tasks. Increasing parameters $s_i$ or $\frac{1}{t}$ adds control text to user input or LLM CoT which does not contribute to task completion.

\section{Further research}
Further research is needed to evaluate Reinforcement in Context against popular alignment benchmarks \cite{harmbench}. Research should include varying parameters $\frac{1}{t}$ and $s_i$ which are the frequency and length of interruptions. Research should also test different prompt texts for the interruptions. These should be context-dependent.

\section{Conclusion}
In this paper we have defined the scaling problem of LLM alignment and presented reinforcement in context through interruptions as solution. Our hope is that research into this subject will contribute to enabling more secure LLM usage, particularly in safety-critical applications.

\section{LLM Usage and Acknowledgment}
Researchers attempted to use generative AI (LLM) for hypothesis validation and to search for examples of reinforcement in context. The LLM did not find examples of similar experiments. We are grateful to the anonymous AI security expert who pointed us to the Anthropic experiment. No part of this paper was LLM-generated.

\printbibliography



%\bibliographystyle{IEEEtran}
\end{document}